# CompNet: A Designated Model to Handle Combinations of Images and Designed features

Bowen Qiu

*DePaul University, Jarvis College of Computing and Digital Media, 243 South Wabash Avenue, Chicago, IL 60604, USA*

Daniela Raicu

*DePaul University, Jarvis College of Computing and Digital Media, 243 South Wabash Avenue, Chicago, IL 60604, USA*

Jacob Furst

*DePaul University, Jarvis College of Computing and Digital Media, 243 South Wabash Avenue, Chicago, IL 60604, USA*

Roselyne Tchoua

*DePaul University, Jarvis College of Computing and Digital Media, 243 South Wabash Avenue, Chicago, IL 60604, USA*

**Abstract**

*Convolutional neural networks (CNNs) are one of the most popular models of Artificial Neural Networks (ANN)s in Computer Vision (CV) for their outstanding ability in handling images. A variety of CNN-based structures were developed by researchers to solve problems like image classification, object detection, and image similarity measurement. Although CNNs have shown their value in most cases, they still have a downside: due to their complexity, they easily overfit when there are not enough samples in the dataset or one class is overrepresented. Most medical image datasets are examples of such a dataset. Additionally, many datasets also contain both designed features and images, but CNNs can only deal with images directly. This represents a missed opportunity to leverage additional information. For this reason – including additional information in an attempt to address the overfitting problem – we propose a new structure of CNN-based model: CompNet, a composite convolutional neural network. This is a specially designed neural network that accepts combinations of images and designed features as input in order to leverage all available information. The novelty of this structure is that it uses learned features from images to weight designed features in order to gain all information from both images and designed features. With the use of this structure on spiculation classification task on LIDC dataset, comparing with using a simple CNN-based model, we show that the training accuracy is decreased from 100% to 92.59% and the testing accuracy is increased from 66.67% to 79.17%. This result indicates that our approach has the capability to significantly reduce overfitting and improve the overall performance when compared to simple CNN-based models. Furthermore, we also found several similar approaches proposed by other researchers that can combine images and designed features. These include the approach that concatenates images and designed features in the middle of the CNN layers when feeding to the model, and the approach that transforms designed features to image-like forms and feed it along with the images to the model. To make comparison, we first applied those similar approaches on LIDC and compared the results with the CompNet results, then we applied our CompNet on the datasets that those similar approaches originally used in their works and compared the results with the results they proposed in their papers. All these comparison results showed that our model outperformed those similar approaches on classification tasks either on LIDC dataset or on their proposed datasets.*

*Keywords:* Convolutional Neural Network, Computer Vision, Overfitting

## 1. Introduction

Convolutional neural networks (CNNs) have evolved into a dramatically increasingly powerful approach among other machine learning methods in computer vision over the past decade because of their outstanding ability in handling images [1]. Researchers can use CNN-based models on various tasks such as classification, object detection, and image similarity measurement. To achieve the designated goal, most state-of-the-art CNN-based models contain a number of stacked layers with abundant parameters to improve the feature extraction capability [2].

While CNNs are ubiquitous, they also have one main disadvantage: CNN-based models easily overfit the data when the dataset is imbalanced and does not have enough samples [3, 4]. The reason is that CNN-based models are usually able to capture large number of features. If there are not many samples in the dataset, then there would not be many features for the model to capture. In this case, the model will over-capture features, which is more likely memorizing every single sample in the training set, and this certainly leads to overfitting. Additionally, CNN-based models only accept images as input; this means that if there are also non-image formed features such as designed features related to the dataset, CNN cannot use them during training and testing. This is a missed opportunity to take advantage of additional available information. If a model can accept both images and designed features at the same time, then theoretically, this model gains the advantage of feature richness when compared to other models that can only accept one kind of feature. Thus, we set out to leverage the combination of different kinds of features in order to improve the lung nodule spiculation prediction accuracy.

This work investigates the overfitting problem and introduces a newly designed structure of CNN-based models that accepts a combination of learned and designed features. To make a comparison, we first run a simple CNN-based model on NIH/NCI's LIDC dataset and use this result as a baseline. The model is overfitted with 100% training accuracy and 66.67% testing accuracy. We then apply our CompNet on the same dataset with designed features combined with images as input. The result ends up with 92.59% training accuracy and 79.17% testing accuracy. We further compare our approach with similar approaches that are proposed by other researchers. We first apply those similar approaches on LIDC dataset and compare the results with both the baseline and CompNet result. The similar approaches' results are better than the baseline, but worse than CompNet result. We then apply our CompNet on the datasets that other researchers used in their own work and compare the CompNet result with their proposed result in their work. The result produced by CompNet is always better with lower training accuracy and higher testing accuracy. Experimental details are shown in later sections. Through this comparison, we show that our approach outperforms. Other similar approaches on LIDC dataset and the datasets they use in their own work.

## 2. Related Works

Images are usually considered as unstructured data and treated differently from structured data in most cases. Traditional machine learning models, such as decision trees, are very good at handling structured data, but cannot deal with images because unlike the values in structured data, a single pixel contains very little meaning on its own. Similarly, deep learning models can deal with images very well, but since structured data usually do not have as many features as unstructured data do, the performance of deep learning models on structured data could be very poor due to overfitting problem.

Ensemble modeling is one solution to solve this problem by training multiple models, some with designed features, others with learned features in isolation then aggregate votes (individual predictions) for the final classification [5]. However, this approach does not make any connection between the learned and designed features since each model works independently, and one model does not know what is happening in another model. The advantage of feature combination is lost.

Other studies suggested injecting the designed features to a CNN-based model by concatenating the feature vector to the output of a fully connected layer [6, 7, 8]. This is a more promising solution because both the image and the designed features are included in the same model during the training process. However, concatenation should only apply to features that belong to the same domain. For example, both features are designed features or both features are learned features, and more important, all features are within the same scale. This is because after concatenation, the combined features will be treated equally, and if two features are from different domains, for example, height of a person and a pixel from an image, the calculation will become inaccurate and biased. Moreover, if the feature scales between the two features have a large difference, the one with larger scale could dominate the classification result. Additionally, concatenation of features makes the features become a whole vector, and the model will lose the chance to learn from interaction of the features during training process.

Another approach introduced by Sharma, Alok, et al [9] showed that by transforming designed features into image-like forms and stacking them to the images can achieve better classification results when compared to using the designed features directly. Nevertheless, this method uses dimensionality reduction techniques to project features into a 2D plane and some information may be lost during this process. Additionally, if the size of images is much larger than the size of designed features, the image-like forms will be very sparse and can only provide limited information.

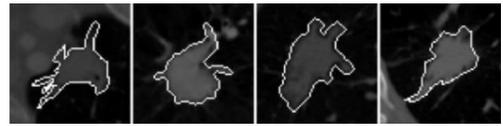

Fig 1: Examples of nodules in LIDC with boundary

So far, to our best knowledge, the most common way to combine images and designed features is using concatenation.

## 3. Methodology

To gain the advantage of feature combination and avoid concatenating features from different domains, our solution is to use the learned features from images as weights of the designed features. The class scores for classification are calculated using these weighted designed features, so the learned features from images (weights) tell the model which designed features are important, and the actual values of the designed features describe how much contribution they can make when the model makes classification. Each sample has its own image and designed features, so the weights learned from one sample could be different from another sample. For this reason, one specific designed feature could have different weights for different samples. Also, the weights are extracted from images but learned based on the importance of designed feature. Thus, the whole training process involves a very strong interaction between images and designed features. The model needs to focus on both sides to learn the relationship between images and designed feature to get properly trained. Comparing with the approaches introduced in [5, 6, 7, 8, 9], this is the main novelty of our approach. To implement this approach, we design a special structure of convolutional neural network: Composite Network (CompNet), and test it on different datasets. In this Section we describe these datasets, followed by our novel CompNet.

### 3.1. The NIH/NCI Lung Image Database Consortium (LIDC)

The Lung Image Database Consortium (LIDC) dataset [10, 11] is the first dataset that we used to compare our approach with previous similar approaches. It contains 2,680 distinct nodules in Computed Tomography (CT) scans from 1,010 patients; nodules of three millimeters or larger are manually identified, delineated, and semantically characterized by up to four different radiologists across nine semantic characteristics, including spiculation and malignancy. Fig. 1 shows four examples of nodules. Because the radiologists may give different ratings to the same nodule, we used the mode of the spiculation ratings as the reference truth.

Besides the nodule images, LIDC dataset also contains 64-dimensional low-level designed features [12]. These features are designed and include shape, texture, intensity, and size related information, such as area, elongation, and extent. The designed features were used to combine with the nodule images in our experiment.

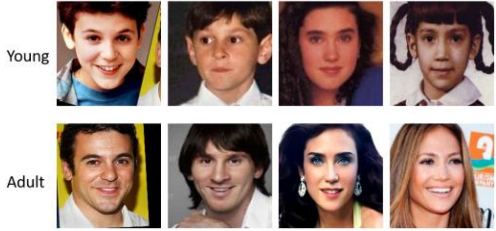

*Fig 2: Sample images of 4 people in LAG. The images in the same column are of the same person*

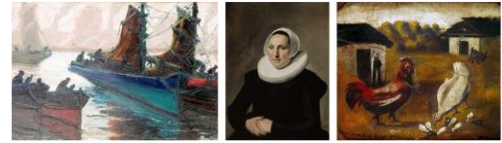

*Fig 3: Sample paintings in MultitaskPainting100k*

### 3.2. The Large Age-gap (LAG)

The Large Age-gap (LAG) dataset was introduced by Bianco et al [6]. It contains 3,831 face images of 1,012 people. Each person has at least one image at a young age and one image at an adult age. Fig. 2 shows eight sample images of four people. In the original work, Bianco's team created 5,051 matching pairs and 5,051 non-matching pairs to do the experiment. They also calculated seven similarity scores, which includes Euclidean Distance, Similarity Metric Learning [16], One Shot Similarity [17], etc. [18, 19], as designed features for each pair using seven different methods.

To compare our approach with Bianco's, we first reproduced Bianco's experiment on LAG dataset, then ran our approach on LAG dataset. All parameters, such as number of pairs, were kept the same.

### 3.3. MultitaskPainting100k

The MultitaskPainting100k dataset was introduced in [7]. This dataset was also used by Bianco et al [7]. MultitaskPainting100k contains 103,250 paintings of 1,508 artists, 125 styles, and 41 genres. Fig. 3 shows three examples in MultitaskPainting100k. The original task proposed in their work was to classify each painting's artist, style, and genre simultaneously. To boost the performance, Bianco et al generated 3,477 designed features using 11 feature descriptors for each painting. These designed features include gray-scale histogram [20], RGB histogram [21], Dual Tree Complex Wavelet Transform [22], etc. [23, 24, 25, 26, 27, 28, 29, 30].

In our experiment, we also calculated those designed features based on the description to reproduce their experiment by our own, and applied our approach on MultitaskPainting100k dataset.

### 3.4. RNA-seq, Vowels, Madelon, and Ringnorm-DELVE

There are 4 datasets introduced in [9]. They are RNA-seq, Vowels, Madelon, and Ringnorm-DELVE. The details of these 4 datasets are shown as following.

RNA-seq dataset is a dataset that contains RNA sequences representing different cancer types. This dataset contains 6216 samples, 60483 features, and 10 classes. Each class is a type of cancer.

Vowels dataset is a speech dataset that contains pure vowels extracted from the TIMIT corpus [13]. In the preprocessing step, each vowel was subdivided into three segments to generate feature vectors so that the dataset became designed. The preprocessed dataset contains 12579 samples, 39 features, and 10 classes. Each sample represents a vowel.

Madelon dataset is a multivariate and highly non-linear artificial dataset that contains 2600 samples, 500 features, and 2 classes [14]. This dataset is to study classification of random data, so this dataset is generated with randomness based on several hyperparameters, which include Gaussian clusters to draw independent features for each class, covariance added based on a random matrix with uniformly distributed, random noise and probes, etc.

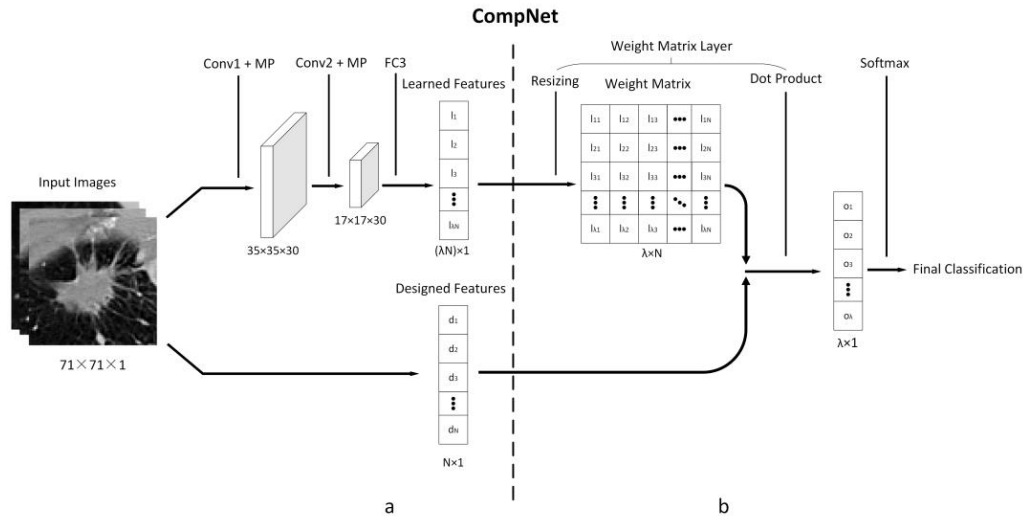

*Fig 4: An overview of CompNet applied on LIDC dataset. The structures used on other datasets are in similar design*

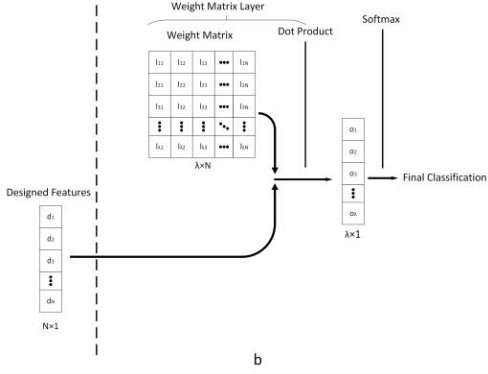

Fig 5: Calculation of class scores

Ringnorm-DELVE is another artificial dataset that contains 7400 samples, 20 features, and 2 classes [15]. Each class is drawn from a multivariate normal distribution. Specifically, class 1 has mean zero and covariance matrix 4 times the identity matrix, and class 2 has unit covariance. Breiman reported that the theoretical expected misclassification rate as 1.3%, but found an error of 21.4% in actual implementation [15].

Unlike the datasets introduced in previous sections, all these four datasets are composed of designed data instead of images. This is because the work that Sharma et al [9]. have done is to transform non-image forms to image-like forms in order to use the power of CNN-based model. We also applied our approach on these 4 datasets to compare the performance with Sharma's work.

### 3.5. Composite Convolutional Neural Network (CompNet)

The Composite convolutional neural network (CompNet) is the proposed network structure in this paper. It is a specially designed CNN that accepts a combination of images and designed features as input. As Fig. 4 shows, there are two parts in CompNet. Part (a) is similar to a traditional CNN. It contains convolutional layers, maxpooling layers, and fully connected layers. Images will be input to part (a) during training and a one-dimensional feature vector $L \in \mathbb{R}^m$, where m is the size of the vector, will be produced as the output. In equation (1), $f$ is a non-linear transformation function that stands for all the layers in part (a). This means any proper input fed to $f$ will be transferred into a one-dimensional feature vector $L \in \mathbb{R}^m$. $I$ stands for the input image, and $L$ stands for the one-dimensional feature vector output mentioned before.

$$L = f(I) \quad (1)$$

Part (b) illustrates the novelty in our approach. The output of part (a), $L \in \mathbb{R}^m$, will be resized to $L` \in \mathbb{R}^{\lambda \times N}$ where $\lambda \times N = m$ while $\lambda$ is the number of classes and $N$ is the number of designed features. To make things clear, we call $L`$ the "*weight matrix*", and the layer producing the weight matrix the "*weight matrix layer*". Equation (2) shows how $L`$ is calculated from $L$.

$$L` = f_{resize}(L) \quad (2)$$

After the *weight matrix* is generated, designed feature vector $D \in \mathbb{R}^N$ is inserted to the model in part (b). A dot product will be applied between the *weight matrix* and the designed feature vector $D$. The result of it, $O \in \mathbb{R}^\lambda$, will be used for softmax classification. Fig. 5 and Equation (3) shows how each $O_k$, where $k \in [1, 2, ... \lambda]$, is calculated.

$$O_k = L_k` \cdot D \quad (3)$$

Unlike the methods in [6, 7, 8], in which the designed features are concatenated, the CompNet treats the learned features from images from part (a), which is $L$ in this case, as weights of the designed features. The novelty of CompNet lies in its learning weights for designed features and training the model using interaction of different features, rather than an abstract embedding for an image. This allows CompNet to learn from the image what the appropriate amount of each designed feature is, and provides a natural importance value for each designed feature. These importance values can then be used to refine the model by removing little used features, if necessary. This is how the model learns from the interaction of learned and designed features.

## 4. Results

### 4.1. LIDC Results

The model presented in Fig. 4 is the general form of our CompNet. In the specifical implementation on the LIDC dataset, $\lambda = 2$ because we only classify if a nodule is spiculated or non-spiculated, which is a 2-class classification problem, and $N = 64$ because each sample has 64 designed features. The size of the input nodule image is 71 by 71 by 1. Both convolutional layers have 30 convolutional kernels (filters). The reason why we picked 30 as the number of filters is that after repeated experiments, we found that 30 is the most optimized number of filters on LIDC dataset. Each convolutional layer is followed by a maxpooling layer respectively. Here we chose LeakyReLU as the activation function to avoid all activations becoming zeros. Since $N = 64$, the size of the output from part (a) was set to be 128 and later resized to 2 by 64. The training curve is shown in Fig. 6. After 2000 epochs, the training accuracy is 92.59% and the testing accuracy is 79.17%.

To compare our result to Bianco and Efthymiou's works [6, 7, 8], we implemented the approaches they proposed in their work, but changed the CNN part to match the part (a) in Fig. 4 so that the results are comparable.

In the original experiment design of Bianco's work [6], the process has 3 parts, which is training part, fine-tuning part, and testing part. In the first part, which is the training part, the author used a simple CNN-based model to train with the data for classification. In this step, the model only uses images and no designed features involved. The second part happens after

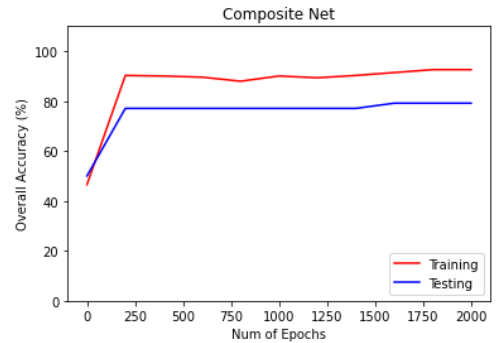

Fig 6: Model performance using approach of CompNet on LIDC in 2000 epochs of training

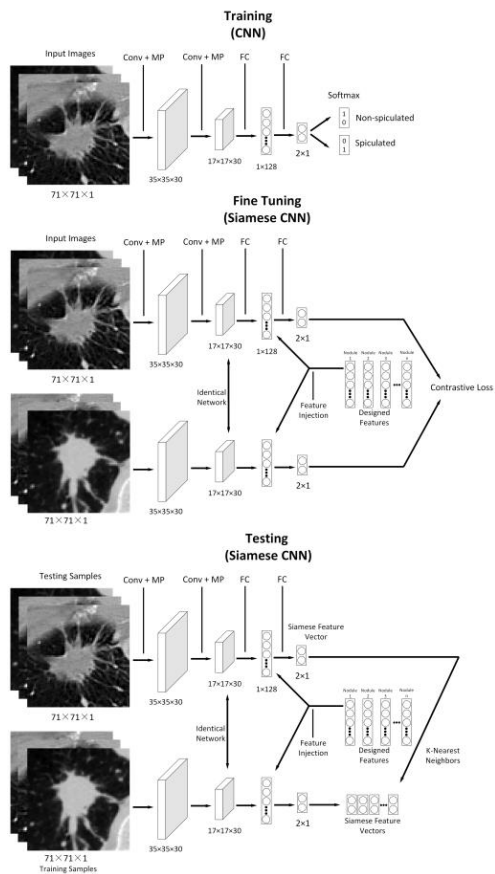

*Fig 7: Implementation of Bianco's approach [6] on LIDC. There are 3 parts in the original approach: Training part, fine tuning part, and testing part. The numbers of convolutional layers and parameters are the same as in CompNet for a better comparison.*

the model is properly trained. In this step, the main task is to fine-tune the model using a Siamese network with identical layers to the previous simple CNN. Pairs of images are used as input to learn similarity and dissimilarity of the images. To be specific, there are two kinds of pairs used. The first one is called positive pairs, which are composed of images of the same person in different ages, and the second one is called negative pairs, which are composed of images of different people. In the middle of training, designed features are integrated to the model and are concatenated to the learned image feature produced by the convolutional layers. Contrastive Loss is the loss function originally used in Bianco's work [6], so we kept the same loss function when applying his approach on LIDC dataset. The purpose of this step is to learn similarity between images of the same person and dissimilarity between images of different people. This is a very common step when dealing with face-recognition tasks. When the parameters of the Siamese network are fine-tuned, the process moves to the third part. This part is the testing process, and pairs of images are fed to the Siamese network to classify if they have the same class label or not. To reproduce this whole experiment on LIDC dataset, we also split the process into 3 parts as described above and follow the same steps. Fig. 7 shows the details of how we reproduced the experiment. We modified the CNN part to match part (a) in Fig. 4, so that the features extracted by the convolutional layers in the reproduced experiment are similar to the features extracted by our approach. We also trained the model in the first and second part for 2000 epochs because we did the same using

CompNet. In this case, the features extracted by the convolutional layers are even more close. After the model is fully trained and fine-tuned in the first 2 parts, we ran the third part for testing. Fig. 8 shows the performance of the first part (upper, CNN Net) and the third part (lower, Concatenation Siamese Net). The training accuracy of the first part is 100% and the testing accuracy is 64.58%. It is obviously overfitting. As discussed earlier, CNN models get overfitted easily if the samples are not sufficient. Since the first part is a CNN model, the performance is expected. For the third part, the training accuracy is 58% and the testing accuracy is 52%.

In Bianco's other work [7], the original experiment design has three branches for three different classification tasks. Since we only have one classification task, which is to classify spiculation, we don't need three branches. Thus, we again modified the CNN part to match part (a) in Fig. 4. The implementation of Bianco's approach [7] on LIDC is shown in Fig. 9, which we built a CNN model with a feature injection layer between the first and second fully connected layer. The images will first go through the convolutional layers and the first fully connected layer like a traditional CNN model does. Then the designed features will be injected to the model by concatenating them to the learned features produced from the first fully connected layer during training and testing. The batch size we used is 64, and the number of epochs is 2000 to match our approach. After the model has completed training, we used the same model to perform testing process. Designed features are used in this step like in training. The performance is shown in Fig. 10. The training accuracy is 100%, and the testing accuracy is 68.75%. The model is obviously overfitted.

In Efthymiou's work [8], the idea of combining images and designed features is to concatenate them together, which is the same as in Bianco's work [7]. In this case, the implementation would be the same and the results are redundant.

The approach provided by Sharma et al [9] presented a different solution. This approach is suggesting to transform designed features from non-image forms to image-like forms in order to use the power of CNN-based models. In this approach, the authors first applied t-SNE technique to reduce the number of dimensions of the features to 2 so that all features are projected into a 2-dimensional space. Based on the mechanism of t-SNE, similar features are close to each other and dissimilar features are far away from each other. This step decides where each feature should go when transforming to the image-like form. After projecting, a convex hull algorithm is used to get the smallest area that can contain all the features. A rotation follows so that the projection is in the same form of an image. Once the projection is completed, the actual feature values fill to the corresponding spots to produce the final samples of image-like forms of designed features. To reproduce this approach on LIDC, we used the same steps to transform designed features of each sample. Fig. 11 shows the projection of each designed feature in LIDC before convex hull algorithm and rotation. We can see that the features are split into two groups. This means that t-SNE technique thinks the features on the left side are similar to each other and features on the right side are similar to each other. When they go through convolutional layers, similar features will be covered by the filter at the same time. However, the authors only had designed features, so they only used transformed image-like designed features to feed the CNN model for training and testing, but on our end, we not only have designed features, but also have images, so we used both transformed designed features and images, and concatenate them together into a 2-channel image-like form (1 channel for nodule image and 1 channel for transformed designed feature). Fig. 12 demonstrates how we implemented Sharma's approach [9]. The upper part is data preprocessing, which transform the designed features to image-like forms and concatenate them to the corresponding image.

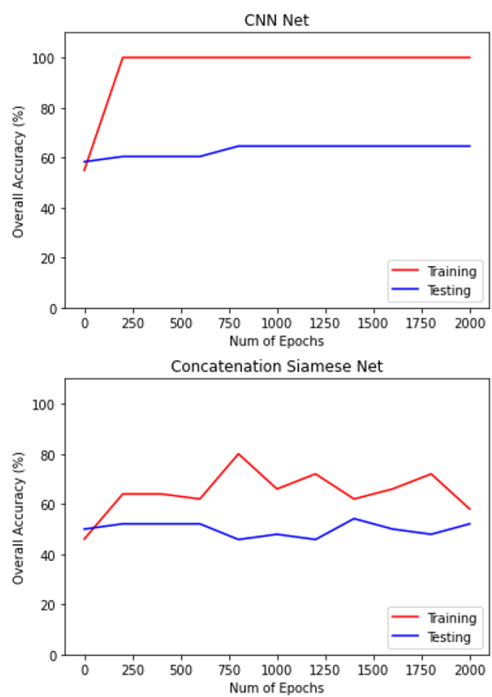

*Fig 8: Model performance using approach of [6] on LIDC. Upper: part 1. Lower: part 3*

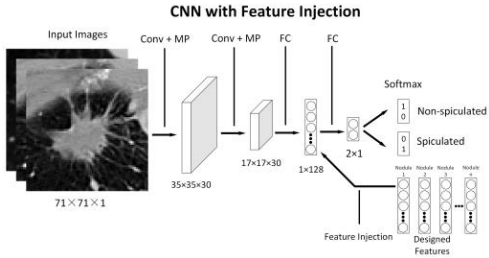

*Fig 9: Implementation of Bianco's approach [7] on LIDC. Between the first and second fully connected layer, we added a feature injection layer, which injects the designed features to the model by concatenating them to the learned features.*

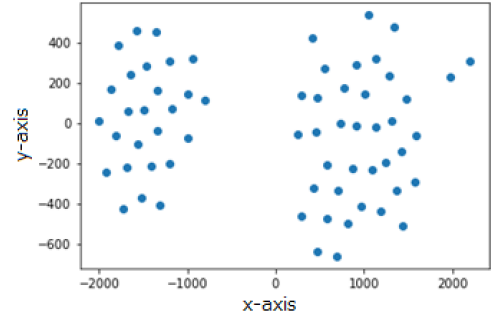

*Fig 11: Raw projection of designed features in LIDC using t-SNE. T-SNE technique thinks features on the left side are similar to each other, and features on the right are similar to each other.*

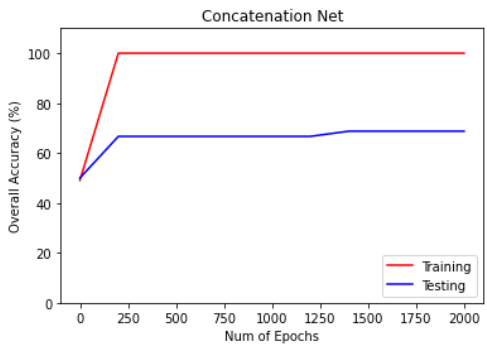

*Fig 10: Model performance using Bianco's approach [7] on LIDC*

The lower part is modeling, which is a CNN model. To run the experiment, we simply fed the concatenated forms to the CNN model. As Fig. 13 shows, the training accuracy is 100% and the testing accuracy is 66.67% indicating overfitting.

### 4.2. LAG Results

We first reproduced Bianco's experiment [6] on LAG. To implement it, we rebuilt the 3 parts of the experiment as discussed in the previous session and applied it on LAG dataset. For the first part, we fed all images to a CNN model for training for 200 epochs. The training accuracy and testing accuracy were 90.21% and 76.40% respectively. Then we created 5051 positive pairs and 5051 negative pairs, and calculated 8 different similarity scores as designed features based on the description of Bianco's work [6] among each pair for the second part implementation. The similarity scores include Euclidean Distance, Similarity Metric Learning [16], One Shot Similarity [17], etc. [18, 19]. The second part is fine-tuning, so we fed all pairs to the Siamese network and injected the similarity scores after the first fully connected layer by concatenating it to the output of the previous layer. The second part also ran for 200 epochs. For the third part, which is the testing part, we used the same network from the second part to predict the pairs in testing set. Similarity scores were involved in this part. After we reproduced the original experiment, we implemented our CompNet on LAG. To make the results comparable, we followed the same steps as we reproduced the original experiment of Bianco's work [6]. We also built 3 parts, and each part was doing the same thing as in the original experiment. The difference is that, in the second and third part when injecting the similarity scores, we replaced the first fully connected layer with a weight matrix layer and kept the rest the same. The training accuracy is 89.30% and the testing accuracy is 81.81%. For our approach, the training accuracy is 87.31% and the testing accuracy is 82.41%.

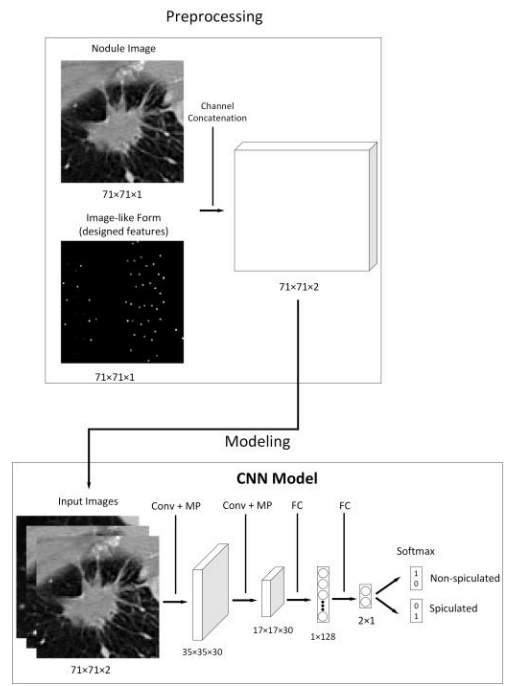

*Fig 12: Implementation of Sharma's approach [9] on LIDC. The upper part is data preprocessing step, which transforms designed features to image-like forms and concatenate it to the corresponding image. The lower part is modeling, which feed the input image to the model.*

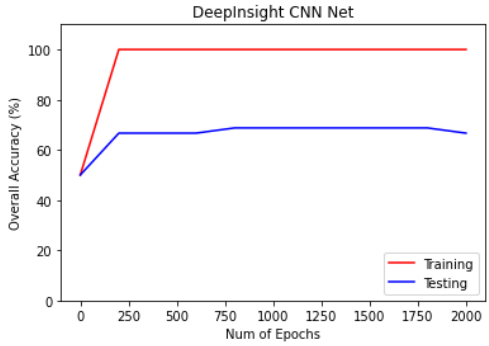

*Fig 13: Model performance using approach of [8] on LIDC*

### 4.3. MultitaskPainting100k Results

The process is similar to what we did in the previous section. In the original experiment of Bianco's other work [7], the authors calculated 11 different designed features and used 3 branches to train and predict artist, tyle, and genre respectively. The 11 designed features include gray-scale histogram [20], RGB histogram [21], Dual Tree Complex Wavelet Transform [22], etc. [23, 24, 25, 26, 27, 28, 29, 30]. To reproduced Bianco's results [7], we also calculated the same designed features and used 3 branches for the same purpose. During training and testing, for each input sample, we extracted 3 ROIs, and fed them to each branch respectively as described in Bianco's work [7]. The hand-crafted features are injected after the branches using concatenation method. After the feature injection, 3 residual blocks followed, and an average pooling was used at the last residual block. 3 parallel fully connected layers are connecting to the residual blocks for the 3 classification tasks. The training process ran for 200 epochs. After the experiment was reproduced, we implemented our approach on MultitaskPainting100k dataset. Most parts of our implementation were the same as when reproducing the experiment, except that we replaced the feature injection part with a weight matrix layer so that the model learns the weight for the designed features when injecting them. The training accuracy is 77.63% (Artist), 91.95% (Style), and 80.49% (Genre), and the testing accuracy is 56.28% (Artist), 54.28% (Style), and 62.61% (Genre). For our approach, the training accuracy is 70.47% (Artist), 83.36% (Style), and 66.61% (Genre), and the testing accuracy is 59.61% (Artist), 56.95% (Style), and 66.61% (Genre).

### 4.4. RNA-seq, Vowels, Madelon, and Ringnorm-DELVE Results

Sharma et al [9] suggested to turn the designed features into image-like forms in order to use the power of CNN models. To do so, the designed features will be first applied with t-SNE to reduce the number of dimensions to 2, so that the features are projected on a 2-dimensional plane, which can be treated as images. After the use of t-SNE, convex hull algorithm, rotation, and resizing are used to tune the output and optimize the performance. The final output image-like forms of each sample will be fed to a CNN model for classification tasks. To reproduce the experiment on each dataset, we followed the same steps mentioned above, transformed the designed features to image-like forms, and fed them to a CNN model. This part was straight forward because it did not have a lot of details in the model structure. The training process ran for 200 epochs for each dataset. To implement our approach on these datasets, we first followed the steps that transform the designed features to image-like forms. Then, we fed the image-like forms to a similar CNN model as we reproduced the experiment. The difference is that the CNN model we used to implement our approach had a weight matrix layer which accepted the original designed features and used the learned features from the image-like forms as weight for the original designed features. The training process had also run for 200 epochs to match the reproduced experiment. The reproduced results for RNA-seq, Vowels, Madelon, and Ringnorm-DELVE are 98.90%, 96.30%, 84.67%, and 98.70% for training, 97.27%, 94.25%, 81.01%, and 98.21% for testing. The results using our approach are 98.47%, 96.94%, 84.81%, and 99.11% for training, 98.10%, 93.26%, 82.09%, and 97.40% for testing.

### 4.5. Result Evaluation

Table 1 shows the summary of results on LIDC dataset. These results are already introduced in previous sections. The column "Our Accuracy" is the testing accuracy on LIDC using CompNet, and the column "Accuracy Using Other Approaches" is showing testing accuracy on LIDC using other similar approaches mentioned before. The rightmost column "Improvement" is showing how much improvement our CompNet made comparing with other similar approaches. Clearly, our approach is better than any other approaches listed in the table. This shows the outstanding ability of our CompNet.

*Table 1: Evaluation on LIDC*

| Related work | Our Accuracy | Accuracy Using Other Approaches | Improvement |
|---|---|---|---|
| Bianco, Simone, 2017 [6] | 79.17% | 52% | +27.17% |
| Bianco, Simone, 2019 [7] | 79.17% | 68.75% | +10.42% |
| Efthymiou, Athanasios, 2021 [8] | 79.17% | 68.75% | +10.42% |
| Sharma, Alok, 2019 [9] | 79.17% | 66.67 % | +12.5% |

To explain why our results produced by the CompNet outperformed Bianco's and Efthymiou's works [6, 7, 8], one possibility would be that, our CompNet focuses more on the interaction of features in the weight matrix layer as shown in Fig. 4 and forces the model to use the relationship between the learned features and designed features by learning weights for designed features while concatenation of features makes the model only train on image embeddings during training. For this reason, our approach can extract extra information when building connections between features and gain a better performance.

When comparing CompNet with Sharma's approach [9] on LIDC, our approach is still better. A good explanation is that, the image-like forms of designed features contributes very little during training and testing, so the performance is almost the same as when feeding with nodule images only without image-like forms of designed features. We already saw how bad when feeding images only, so this result is expected

Table 2 shows the summary of results when applying on other datasets than LIDC. Similar to table 1, the column "Our Accuracy" shows the testing accuracy of CompNet applied on different datasets. Though some performances are not very good, for example, on MultitaskPainting100k, but overall, there are improvement using our CompNet comparing with the reproduced results using their own approaches.

*Table 2: Evaluation on Other Datasets*

| Dataset | Our Accuracy | Accuracy Using Their Own Approaches | Improvement |
|---|---|---|---|
| LAG [6] | 82.41% | 81.81% | +0.60% |
| MultitaskPainting100k [7] | 59.46% (Artist) 56.95% (Style) 66.37% (Genre) | 56.28% (Artist) 54.28% (Style) 62.61% (Genre) | +3.18% +2.67% +3.76% |
| RNA-seq [9] | 98.10% | 97.27% | +0.83% |
| Vowels [13] | 93.26% | 94.25% | -0.99% |
| Madelon [14] | 82.09% | 81.01% | +1.08% |
| Ringnorm-DELVE [15] | 97.40% | 98.21% | -0.81% |

## 5. Conclusion and Future Work

In this work we propose a new CNN-based structure, the CompNet, which leverages both learned features and designed information about the images, reducing overfitting, a known problem for CNNs. We show that our proposed approach not only reduces overfitting but also slightly improves the overall performance of the network in some cases when tested against other approaches and other datasets. Instead of concatenating the features together, our approach used a dot product to emphasize the interaction between the different types of features. This is the key contribution of the CompNet responsible for the improvement over previous concatenation-only methods.

Additionally, because we created a weight vector in the model (treated as weights of designed features), we can further explore the relationship between the class labels and each designed feature by analyzing the values in weight vector.